\documentclass[11pt,a4paper]{article}
\usepackage{times,latexsym}
\usepackage{url}
\usepackage[T1]{fontenc}

\usepackage[acceptedWithA]{tacl2021v1}

\usepackage{xspace,mfirstuc,tabulary}

\newif\iftaclinstructions
\taclinstructionsfalse 
\iftaclinstructions
\renewcommand{\confidential}{}
\renewcommand{\anonsubtext}{(No author info supplied here, for consistency with
TACL-submission anonymization requirements)}
\newcommand{\instr}
\fi

\iftaclpubformat 

\else

\fi

\date{}

\usepackage{times}
\usepackage{latexsym}
\usepackage{amssymb}
\usepackage{stix}
\usepackage{graphicx}
\usepackage{subfig}
\usepackage{booktabs} 
\usepackage{multirow}
\usepackage{tikz}

\usepackage[T1]{fontenc}

\usepackage[utf8]{inputenc}

\usepackage{microtype}

\usepackage{inconsolata}

\title{Hypernetworks for Personalizing ASR to Atypical Speech}

\author{
  Max M\"uller-Eberstein$^{1+*}$ \
  Dianna Yee$^{2*}$   Karren Yang$^2$   Gautam Varma Mantena$^2$   Colin Lea$^2$
  \\
  \ \\
  $^1$IT University of Copenhagen
  \\
  \texttt{mamy@{itu.dk}}
  \\
  \\
  $^2$Apple
  \\
  \texttt{\{dianna\_yee, karren\_yang, gmantena, colin\_lea\}@{apple}.com}
}

\begin{document}
\maketitle
\def\thefootnote{$*$}\footnotetext{These authors contributed equally to this work.}\def\thefootnote{\arabic{footnote}}
\def\thefootnote{$+$}\footnotetext{Research performed while at Apple.}\def\thefootnote{\arabic{footnote}}

\begin{abstract}
Parameter-efficient fine-tuning (PEFT) for personalizing automatic speech recognition (ASR) has recently shown promise for adapting general population models to atypical speech. However, these approaches assume a priori knowledge of the atypical speech disorder being adapted for---the diagnosis of which requires expert knowledge that is not always available. Even given this knowledge, data scarcity and high inter/intra-speaker variability further limit the effectiveness of traditional fine-tuning. To circumvent these challenges, we first identify the minimal set of model parameters required for ASR adaptation. Our analysis of each individual parameter's effect on adaptation performance allows us to reduce Word Error Rate (WER) by half while adapting 0.03\% of all weights. Alleviating the need for cohort-specific models, we next propose the novel use of a meta-learned hypernetwork to generate highly individualized, utterance-level adaptations on-the-fly for a diverse set of atypical speech characteristics. Evaluating adaptation at the global, cohort and individual-level, we show that hypernetworks generalize better to out-of-distribution speakers, while maintaining an overall relative WER reduction of 75.2\% using 0.1\% of the full parameter budget.
\end{abstract}

\section{Introduction}

Large-scale automatic speech recognition (ASR) models are trained predominately on speech collected from the general population and historically have not been able to fully-support speakers with atypical speech. Recent works have proposed parameter-efficient fine-tuning (PEFT) of large ASR models for adapting such general population models to work better for people with speech differences \citep{tomanek2021ondevice, tomanek2021adapters, qi2023parameterefficient}. Such adaptations have focused either on fine-tuning using data from a group of individuals with common speech differences---referred to as cohort-level fine-tuning---or on fine-tuning on speech data at the level of an individual.

Individually personalized ASR models yield state-of-the-art transcription performance, however they require laborious data collection, which could be especially strenuous for people with severe speech disorders. Additionally, characteristics vary greatly, even for the same speaker over time, potentially leading to data drift and eventual performance degradation ~\citep{Tomanek2023}. 

At the cohort level, PEFT of general population models has been shown to improve transcription performance for dysarthria \citep{tomanek2021adapters}. While this approach reduces training data and compute requirements, it requires a priori knowledge of an individual's atypical speech category, which is not always available. As we later demonstrate, a precise diagnosis is crucial, as fine-tuning on a specific cohort does not transfer well to other types of atypical speech (Section \ref{sec:transferability}). 
Furthermore, such solutions require discrete categorizations of individuals and do not share knowledge across cohorts, although sharing may be beneficial for individuals who express mixtures of speech differences, or between individuals with different severities of the same speech disorder.

In this work, we consolidate individual-level personalization with knowledge-sharing across cohorts by proposing the use of hypernetworks to generate adaptation parameters dynamically during inference, for individualized personalization amongst a heterogeneous cohort of speech disorders. As opposed
to cohort and individual-level fine-tuning, which learn fixed adaptations that are difficult to transfer across etiologies, hypernetworks leverage a meta-learning procedure that instead learns to generate adaptation parameters, conditioned on the target speaker's speech characteristics. This approach enables flexibility with respect to the adaptations applied to the ASR model, as they can change depending on the individual utterance being adapted for. Simultaneously, the use of a single hypernetwork instead of multiple pre-trained adaptations for each cohort or individual reduces complexity and actively promotes sharing information that is useful across different types of atypical speech.

In our study, we include both phonological and fluency-related speech disorders, using speech from people with \textit{stuttering}, \textit{dysarthria} consistent with Cerebral Palsy, and \textit{Parkinson's} disease, for which a myriad of speech differences including dysarthria and stuttering may be exhibited. Stuttering includes dysfluencies, such as sound, word or phrase repetitions (``m-m-mall'', ``go go go''), prolongations (``baaall''), and audible pauses or blocks \cite{sander1963frequency, SSI}. Dysarthric speech may contain differences in pronunciation, pitch, intelligibility, strain, speaking rate and volume. It is particularly challenging as the expressed characteristics depend on the etiology and individual, varying even for one speaker~\citep{rowe2022}. For example, spastic dysarthria, commonly associated with Cerebral Palsy, is characterised by slow speaking rates, strained voice and pitch breaks~\citep{theresa2013dysarthria}, whereas hypokinetic dysarthria, commonly associated with Parkinson's disease, is characterized by monotonous speech, varying in volume, breathiness, hoarseness, rapid repetition of phones and imprecise consonant production \citep{duffy1995motor,tjaden2008dysarthria}.

Towards improving ASR for these speech communities, we concretely contribute: 
\begin{itemize}
    \item To the best of our knowledge, the first study of adapting transformer-based ASR models to dysarthric, dysfluent and Parkinson's-influenced speech simultaneously; 
    \item The highest resolution analysis to date, regarding which model parameters contribute most to adaptation (Section \ref{sec:cohort-personalization}); 
    \item A novel approach of using hypernetworks to generate individualized, zero-shot ASR adaptations dynamically across atypical speech types (Section \ref{sec:dynamic-personalization}); 
    \item Experiments covering global, cohort and individual adaptation, to compare hypernetworks with prior work and analyze factors important to its performance (Section \ref{sec:results}).
\end{itemize}

\section{Related Work}

Personalized ASR adaptation for atypical speech is a broad yet under-explored topic. Prior works mainly focus on large-scale ASR models trained on general population speech, which are subsequently fine-tuned on small datasets of atypical speech \citep{shor2019personalizing, green2021automatic}. Such datasets are scarce, and even more so at the level of individual speakers, leading to over-fitting and poor generalization. To overcome these challenges, past works have explored PEFT methods such as individually re-weighting transcription output probabilities \citep{morales2007confusion}, or using residual adapters \citep{Rebuffi_2017} to individually personalize ASR models ~\citep{tomanek2021adapters}, while retaining the original model weights and only training the light-weight adapter modules.

Another approach leverages cohort-level transfer learning: \citet{tomanek2021ondevice} use a two stage fine-tuning process, where the ASR model is first fine-tuned on data from a cohort sharing atypical speech characteristics, before being further fine-tuned on data from an individual in the same cohort. \citet{qi2023parameterefficient} follow a similar approach, but make use of less resource-intensive adapter fusion \citep{pfeiffer2021adapterfusion}, where a cohort-level adapted model is fused with multiple individual-level adapters to train personalized models for new target speakers.

These aforementioned works either require sufficient data from the target speaker in order to train a personalized model, and/or knowledge of the cohort an individual belongs to---both of which may not be readily available. As we demonstrate in Section \ref{sec:transferability}, the process of maintaining and selecting the correct cohort model is critical, however defining the cohort is nontrivial as assigning membership may not be limited to etiology but also severity thereof. Furthermore, prior approaches consider cohorts independently of each other and are thus unable to share knowledge that may be beneficial for better generalization performance across individuals.

In order to generate individualized adaptations while learning globally shared representations across cohorts, we reformulate ASR adaptation as a meta-learning problem. We propose to model this inductive bias via a light-weight hypernetwork meta-learner \citep{ha2017hypernetworks}, which is tasked to generate the most effective adaptation weights for an individual based on their speaker characteristics as represented by a shared encoder.

While this work, to the best of our knowledge, is the first to apply hypernetworks to ASR, recent works have applied them to predict the task-specific adaptations of a text-based, pre-trained, large-scale Transformer architecture ~\citep{mahabadi2021parameterefficient, phang2023hypertuning}. Additionally, language model adaptations generated by hypernetworks have also been shown to generalize to unseen task and language  combinations~\cite{ ansell-2021-mad-g, ustun2022udapter, üstün2022hyperx}. Based on these results as well as recent successes in adapter fusion \citep{pfeiffer2021adapterfusion, qi2023parameterefficient}, we hypothesize that zero-shot personalization is possible by having the hypernetwork learn a mapping between speaker characteristics and ASR adaptation weights, effectively learning a manifold of personalized models. This procedure would require neither labelled audio data nor fine-tuning on the individual-level, leading to increased parameter and data efficiency.

\section{Setup}
\label{sec:setup}

\subsection{Data}
\label{sec:dataset}
Our experiments use three datasets containing speech with phonological and fluency-related speech disorders, with content relating to common voice commands for digital assistants, as well as dictation.
The first dataset $X_{D}$, as described in~\citet{lea2023latent}, contains dysarthric speech with varying severities mostly consistent with Cerebral Palsy.
All 33 participants read a common set of 51 phrases with at least 5 repetitions in multiple recording sessions with several microphone placements, across multiple days.
The second dataset $X_{S}$, as described in~\citet{lea2023stuttering}, contains speech from people who stutter with various degrees of fluency.
All 91 participants were prompted from a common set of dictation and voice assistant tasks but had agency to personalize the commands.
The speech of all participants within $X_{D}$ and $X_{S}$ was graded by a Speech-Language Pathologist as `mild', `moderate', or `severe'.
The third dataset $X_{P}$, is a subset of the Speech Accessibility Project~\cite{SAProject}, which contains speech from 113 individuals whose speech is consistent with Parkinson's disease, saying a mixture of read and free-spoken prompts for common dictation and voice assistant commands, and has not been graded by severity.
The full public benchmark, denoted by $X_\mathbb{P}$\footnote{Planned for public release in Spring 2024.}, contains a broader set of 253 participants, for which we run additional experiments to provide official benchmark results for future work.

In each setup, we run a preliminary study with 3 random seeds where $X_D$, $X_S$ and $X_P$ are split with no speaker overlap into 70\% train, 10\% validation and 20\% test sets, and there is no known overlap of participants across datasets.

\subsection{Models}

We choose Whisper \citep{radford2023whisper} as our base model architecture, a series of 10 encoder-decoder Transformer models, which vary with respect to model size and pre-training data. Trained on 680k hours of general population speech, they allow us to investigate how well the largest contemporary models for typical speech characteristics fare in low-resource adaptation scenarios. 

To gain an understanding of how personalization affects the model, we ablate across multiple fine-tuning setups (Section \ref{sec:cohort-personalization}). Going beyond previous works, we first run an extensive full fine-tuning sweep, additionally ablating across seven partial model components and layers. Next, we adapt sub-layer components, such as the attention and feed-forward layers, using Low-rank Adaptation (LoRA; \citealp{lora}). We respectively denote these setups as $\{\textit{full}, \textit{partial}, \textit{LoRA} \}$. Based on these ablations, we identify which parameters contribute most to personalization, and then train hypernetworks to generate them dynamically (Section \ref{sec:dynamic-personalization}).

\subsection{Evaluation}
\label{sec:evaluation}
Since our proposed approach aims to generate dynamic personalizations, which generalize across a heterogeneous collection of speech disorders, the hypernetwork is trained using a concatenation of all three atypical speech types $X_D$, $X_S$, $X_P$. In our final experiments, we further include a hypernetwork solely trained on the $X_{\mathbb{P}}$ benchmark to enable future comparisons, as well as to investigate the effects of lower speaker diversity.

As prior works have mainly focused on cohort-level or individualized fine-tuning \citep{tomanek2021adapters, tomanek2021ondevice, qi2023parameterefficient}, we define our baselines correspondingly, i.e., with access to speech disorder diagnoses. Additionally, we re-train these cohort-level baselines using the same concatenated training datasets as the hypernetwork to ensure a fair comparison. These dataset-specific and concatenated setups are denoted respectively by $\{\textit{cohort}, \textit{global}\}$. Finally, we evaluate personalization at the most granular, \textit{individual} level, by continually fine-tuning/adapting the matching cohort-level baselines on training data from the target speaker. Note that, while the baselines assume access to the speaker's cohort information and even individualized training data, our hypernetwork-based approach is the first to operate in a fully zero-shot manner, without additional speaker-specific training (meta-)data. All evaluations are computed using the same sets of utterances, and of speakers not observed during training, using the transcription Word and Match Error Rates (WER, MER; \citealp{morris2004mer}). For the individual-level adaptation experiments, the baselines use part of a target speaker's utterances as training data and use the remainder as unseen test data, while the hypernetwork remains completely agnostic to the target speaker and is applied directly to the equivalent test data subset without additional training.

In our experiments, we observe that Whisper occasionally hallucinates, especially for stuttered speech, repeatedly decoding a stuttered syllable up until its maximum decoding length, even if the audio actually contains further content. While MER normalizes this high number of insertion errors into a [0, 100] range, for WER these hallucinations result in large, anomalous values, which hinder the comparison of results across setups. We therefore report performance in terms of median (P50) and Interquartile Range (IQR) WERs on the speaker level for robustness against such outliers.

\section{Cohort-level Personalization}\label{sec:cohort-personalization}

Generating an entire personalized model would be prohibitively expensive. As such we first follow the cohort-level personalization paradigm to identify which pre-trained models provide good initializations for adaptation (Section \ref{sec:untuned}), and which components (Section \ref{sec:tuned}) and individual parameters (Section \ref{sec:lora}) are crucial to fine-tune and adapt. This allows us to create a lighter-weight adaptation framework to which we later apply dynamic personalization (Section \ref{sec:dynamic-personalization}).

\subsection{Pre-trained Model Performance}\label{sec:untuned}

\begin{figure}
    \includegraphics[width=\linewidth]{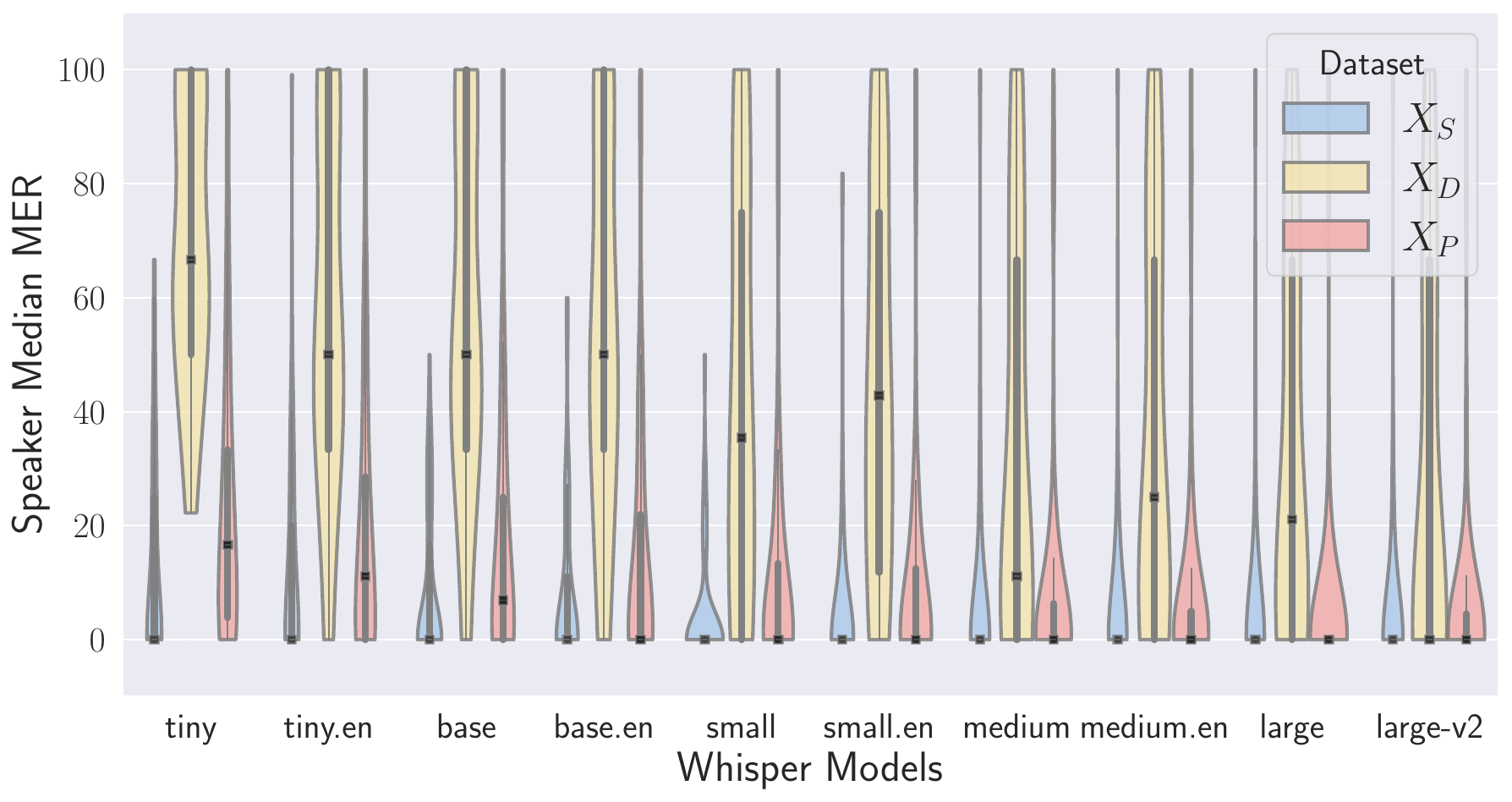}
    \caption{Speaker median MER of untuned pre-trained Whisper models on $X_S$ (stuttering), $X_D$ (dysarthria) and $X_P$ (Parkinson's).}
    \label{fig:untuned}
\end{figure}

\begin{table*}[thb]
    \centering
    \caption{Average speaker WER of untuned, fully tuned, low-rank adapted Whisper models on test splits of $X_P$ (Parkinson's), $X_S$ (stuttering) and $X_D$ (dysarthria), reported as `P50 {\small(IQR)}'. Best fully tuned setups in \textbf{bold}, and best low-rank tuned setups in \textbf{\textit{bolded-italics}}. Setups vary as to whether all components (\textsc{All}), only the encoder (\textsc{Enc}) or decoder (\textsc{Dec}), or the earlier ($\downarrow$) or later ($\uparrow$) layers thereof were trained/adapted. Percentages indicate the amount of tuned parameters with respect to full fine-tuning.}
    \label{tab:ablations}
    \resizebox{\textwidth}{!}{
    \begin{tabular}{@{}p{0.01cm}p{0.001cm}ccccccccccc@{}}
        \toprule
        \multicolumn{2}{c}{\multirow{3}{*}[-2pt]{\textsc{Setup}}} & 
        \textsc{Untuned} & 
        \multicolumn{7}{c}{\textsc{Full Tuning}}& \multicolumn{3}{c}{\textsc{LoRA}}  \\
        \cmidrule(lr){3-3}
        \cmidrule(lr){4-10}
        \cmidrule(lr){11-13}
        &  & - & \textsc{All} & \textsc{Enc} & \textsc{Enc}$_\downarrow$ & \textsc{Enc}$_\uparrow$  & \textsc{Dec} & \textsc{Dec}$_\downarrow$ & \textsc{Dec}$_\uparrow$ & \textsc{All}  & \textsc{Enc} & \textsc{Dec}  \\
        & &  {\small0\%} & {\small100\%} & {\small28\%} & {\small14\%} & {\small14\%} & {\small72\%} & {\small36\%} & {\small36\%} & {\small9\%} & {\small4\%} & {\small6\%}  \\
    \midrule
    \addlinespace[.5em]
        \parbox[t]{2mm}{\multirow{3}{*}{\rotatebox[origin=c]{90}{\footnotesize{\texttt{tiny}} }}}  &
            $X_{P}$ & 25.9 {\small(39.4)} &	\textbf{11.7 {\small(25.9)}} &	22.4 {\small(55.5)} &	28.0 {\small(42.7)} &	25.5 {\small(37.8)} &	23.4 {\small(36.8)} &	25.8 {\small(36.9)} &	25.7 {\small(37.7)} &	\textbf{\textit{24.1 {\small(38.3)}}} &	27.0 {\small(39.5)} &	26.2 {\small(37.8)} \\
           & $X_{S}$ & 16.8 {\small(171.5)} &	\textbf{0.5 (9.2)} &	2.6 {\small(15.1)} &	9.7 {\small(32.0)} &	2.3 {\small(11.9)} &	6.8 {\small(21.2)} &	8.6 {\small(28.0)} &	2.3 {\small(14.0)} &	\textbf{\textit{2.6 {\small(14.7)}}} &	5.8  {\small(22.9)} &	8.7 {\small(28.3)} \\
            & $X_{D}$ & 82.7 {\small(96.8)} &	\textbf{2.6 (9.5)} &	9.0 {\small(16.7)} &	35.7 {\small(62.1)} &	9.1 {\small(16.9)} &	9.5 {\small(20.8)} &	58.1 {\small(81.6)} &	7.3( 14.7) &	\textbf{\textit{19.7 {\small(46.3)}}} &	21.2 {\small(46.5)} &	45.2 {\small(81.1)} \\
        \addlinespace[.5em]
        \midrule
        \addlinespace[.5em]
        \parbox[t]{2mm}{\multirow{3}{*}{\rotatebox[origin=c]{90}{\footnotesize{\texttt{tiny.en}}}}}  & 
            $X_{P}$ & 22.4 {\small(34.1)} &	18.8 {\small(31.3)} &	19.2 {\small(32.4)} &	22.3 {\small(34.0)} &	\textbf{17.7 {\small(30.4)}} &	20.5 {\small(31.5)} &	22.8 {\small(34.0)} &	22.0 {\small(31.7)} &	21.7 {\small(35.2)} &	\textbf{\textit{19.1{\small(31.2)}}} &	22.9 {\small(34.5)}  \\
            & $X_{S}$ & 97.3 {\small(239.2)} &	\textbf{0.2 (7.6)} &	2.0 {\small(12.0)} &	5.1 {\small(19.8)} &	1.5 {\small(10.6)} &	2.7 {\small(11.9)} &	5.0 {\small(17.8)} &	2.8 {\small(12.4)} &	\textbf{\textit{0.3 {\small(8.8)}}} &	2.2 {\small(13.6)} &	3.6 {\small(13.8)} \\
            & $X_{D}$ & 67.4 {\small(91.8)} &	1.9 {\small(9.5)} &	11.1 {\small(18.8)} &	24.6 {\small(48.5)} &	6.4 {\small(12.6)} &	4.8 {\small(11.9)} &	20.6 {\small(42.4)} &	\textbf{0.0 {\small(9.5)}} &	\textbf{\textit{3.6 {\small(9.5)}}} &	13.1 {\small(21.5)} &	7.6{\small(14.2)}\\
        \addlinespace[.5em]
        \midrule
        \addlinespace[.5em]
        \parbox[t]{2mm}{\multirow{3}{*}{\rotatebox[origin=c]{90}{\footnotesize{\texttt{large-v2}}}}}  & 
            $X_{P}$ & 8.5 {\small(17.7)} &	5.2 {\small(9.4)} &	7.4 {\small(15.5)} &	7.5 {\small(15.1)} &	7.5 {\small(15.6)} &	\textbf{5.0 {\small(10.1)}} &	5.5 {\small(13.4)} &	5.8 {\small(11.5)} &	\textbf{\textit{4.9 {\small(10.2)}}} &	5.9 {\small(13.6)} &	5.5 {\small(11.1)} \\
            & $X_{S}$ & 19.3 {\small(171.0)} &	\textbf{0.0 {\small(1.0)}} &	0.0 {\small(3.9)} &	0.0 {\small(5.8)} &	0.0 {\small(4.0)} &	0.0 {\small(1.5)} &	0.1 {\small(3.8)} &	0.0 {\small(2.0)} &	\textbf{\textit{0.0 {\small(2.7)}}} &	1.5 {\small(7.9)} &	0.2 {\small(4.2)} \\
            & $X_{D}$ & 33.3 {\small(63.8)} &	\textbf{0.0 {\small(2.4)}} &	2.8 {\small(6.3)} &	11.3 {\small(16.3)} &	2.3 {\small(6.0)} &	0.0 {\small(5.3)} &	0.0 {\small(9.4)} &	0.0 {\small(4.0)} &\textbf{\textit{0.0 {\small(6.6)}}} &	11.7 {\small(19.2)} &	0.0 {\small(9.5)}   \\
        \addlinespace[.5em]
        \bottomrule
    \end{tabular}
}
\end{table*}

Across the 10 different Whisper model sizes and pre-training paradigms, their transcription error rates on each dataset in Figure \ref{fig:untuned} show that performance tends to improve with model size. However, even the 1.6B parameter \texttt{large-v2} model is not performant enough, presuming a usable system has WER < 15\% \cite{projecteuphonia}. We also observe that Whisper models tend to transcribe verbatim (e.g., stuttered repetitions), which may not be desirable for some downstream applications~\citep{lea2023stuttering}. Additionally, the most severe errors arise from infinite repetition loops in the decoding process, to which the monolingual variants appear to be slightly more robust. Our subsequent experiments focus on both ends of the model spectrum: the multilingual \texttt{tiny} and \texttt{large-v2} models, with 39M and 1.6B parameters respectively, plus the monolingual English \texttt{tiny.en}, also with 39M parameters. 

\subsection{Full Component-level Fine-tuning}\label{sec:tuned}

Full fine-tuning provides a theoretical upper-bound of ASR performance and compute requirements. In addition, we ablate training seven sub-architectures, including the full encoder/decoder, the earlier$_\downarrow$/later$_\uparrow$ half of their respective layers, and the final decoder head. In terms of sub-architectures, Table \ref{tab:ablations} shows that tuning all parameters in either the encoder or decoder leads to the lowest WER in most cases. Notably, \texttt{large-v2} is able to reach median speaker-wise WERs of close to 0, with tight interquartile ranges within the usability threshold of 15\% WER. This indicates that it should theoretically be possible to achieve good coverage of most common voice commands across our examined speech disorders, given sufficient model capacity and compute. Tuning earlier or later parts of the model, such as the early encoder layers and the final decoding head, exhibited higher instability and worse performance. These patterns generalize across the three types of atypical speech as well as across model sizes and multi/monolinguality.

\subsection{Parameter-efficient Sub-layer Adaptation}\label{sec:lora}

Given that the optimal sub-architectures for full fine-tuning generalize well, we aim to localize the optimal sub-layer components for adaptation using LoRA \citep{lora}. In contrast to residual adapters, which adapt only the the Multi-layer Perceptron (MLP) component of each Transformer block, LoRA allows for more targeted adaptation, including the query, value and attention matrices in each layer. To target these individual parameters, LoRA augments any pre-trained weight matrix $W$ by adding a trainable low-rank matrix $\Delta W$. The adapted weight $W'$ is defined by
\begin{equation}\label{eq:lora}
    W' = W +  \Delta W = W + BA \hspace{.2cm},
\end{equation}
where $\Delta W$ is rank $r$ and factorized by two low-rank matrices $A$ and $B$.

We observe that adapting self-attention is unstable and rarely yields performance benefits. In contrast, the highest performance gains stem from adapting all components (Table \ref{tab:ablations}), or the MLPs at the end of each layer. Applying LoRA ($r =$ 64) to the MLPs alone, can thereby reduce training costs down to 4\% of full fine-tuning, while retaining equivalent or better WER. These observations once again hold across all setups.

\subsection{Parameter-level Adaptation Magnitudes} \label{sec:ssa}
\begin{figure}
    \centering
    \subfloat[\centering $X_{D}$]{{\includegraphics[width=0.3\linewidth]{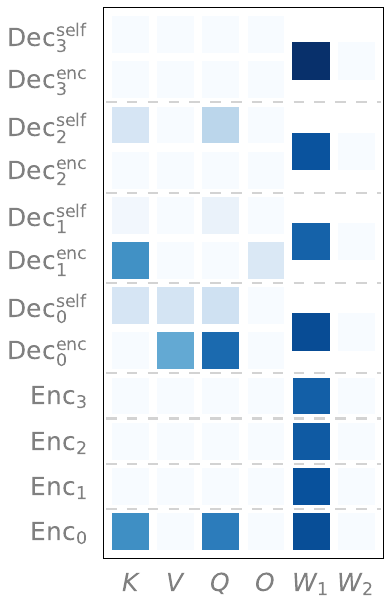} }}%
    \subfloat[\centering $X_{S}$]{{\includegraphics[width=0.3\linewidth]{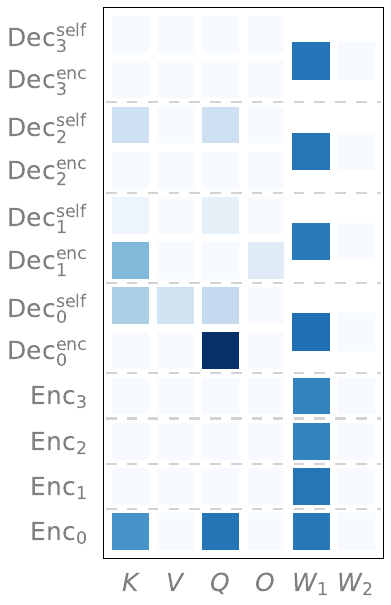} }}%
    \subfloat[\centering $X_{P}$]{{\includegraphics[width=0.3\linewidth]{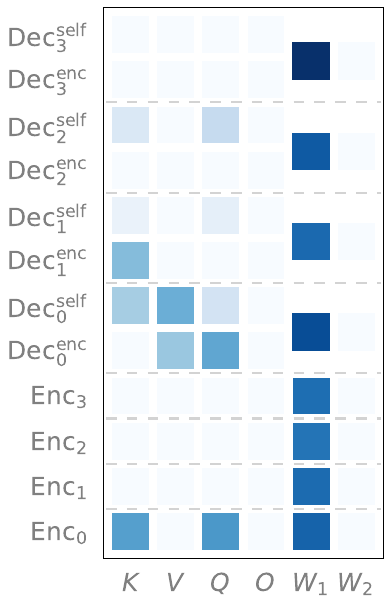} }}%
    \caption{Adaptation magnitudes of key $K$, value $V$, query $Q$, output projection $O$ matrices of the self/cross-attention components, as well as the first $W_1$ and second $W_2$ layers of the MLP within each Whisper (\texttt{tiny.en}) encoder/decoder layer, measured in SSAs according to Section \ref{sec:ssa}.}%
    \label{fig:ssa-heatmap}%
\end{figure}

\begin{figure*}
    \centering
    \includegraphics[width=.93\linewidth]{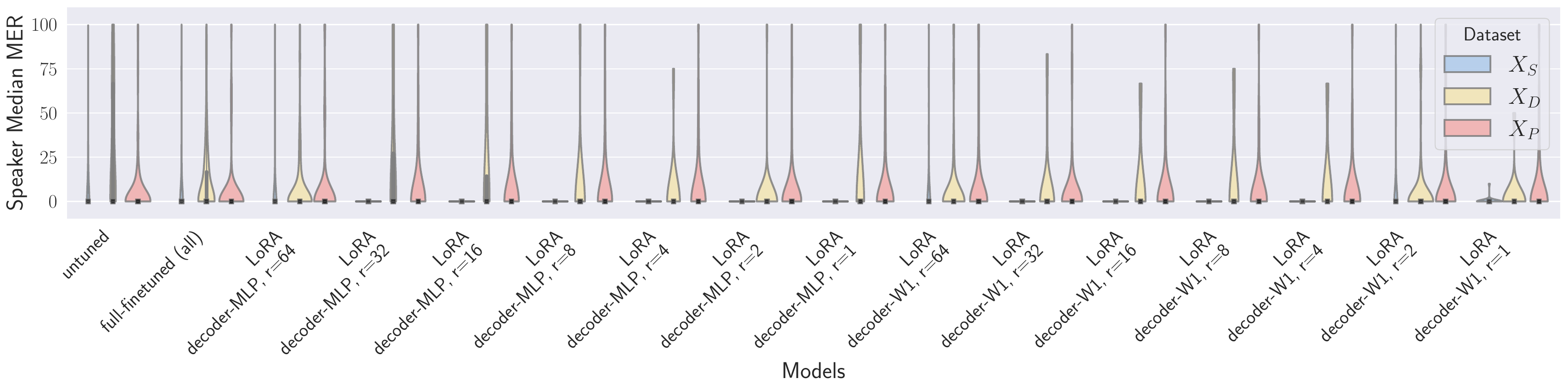}
    \caption{Speaker median MER on $X_S$ (stuttering), $X_D$ (dysarthria) and $X_P$ (Parkinson's) of Whisper (\texttt{large-v2}) untuned, fully-tuned, and adapted using LoRA at both MLP layers or $W_1$, using $r \in [2;64]$.}
    \label{fig:rankablation}
\end{figure*}

To understand the magnitude and localization of adaptations at a higher level of detail---specifically for individual parameters---we propose measuring the difference between each original weight $W$ and its adapted matrix $W'$ using Principal Subspace Angles (SSAs; \citealp{knyazev2002ssa}). This measure keeps adaptation magnitudes comparable across different dimensionalities of $W$ irrespectively of linear invariance by using the singular values of the transformation between the orthonormal bases of the two matrices to measure the ``energy'' required to map one to the other, expressed as an angle from 0$^{\circ}$ to 90$^{\circ}$ (similar/dissimilar).

We compute SSAs at the parameter level, plotting the resulting angles for \texttt{tiny.en} in Figure \ref{fig:ssa-heatmap}.
We observe that the largest adaptation is concentrated in the first linear transformation $W_1$ of the MLP. Some adaptations are learned for the key $K$ and query $Q$ matrices of the early encoder and decoder layers, however these are sparser and less pronounced. This pattern is consistent across all datasets and best-performing model configurations. To confirm these findings, we run experiments where only $W_1$ is adapted and compare the performance to when the entire MLP is adapted in Figure \ref{fig:rankablation}. We observe similar and even improved performance comparable to full fine-tuning in some cases. We thus conclude that $W_1$ is necessary and effective to adapt.

To pursue further parameter efficiency, we consider reducing the rank of the LoRA matrix $\Delta W$, and focus on the decoder specifically, for which we observed larger improvements compared to tuning/adapting only the encoder (Table \ref{tab:ablations}). Similar to observations in \citet{lora}, Figure \ref{fig:rankablation} shows that adaptations are robust to the reduction in rank, down to even rank 2 and 1. By localizing the individual parameter type most relevant to adaptation, we are thus able to effectively halve WER while using 0.03\% of the full parameter budget.

Based on these findings, our subsequent experiments for dynamic personalization via hypernetworks therefore focus on learning to adapt $W_1$ of each MLP in the decoder. Furthermore, each of these $W_1$ matrices will be adapted using LoRA following Equation \ref{eq:lora} with rank $r=2$.

\subsection{Transferability across Etiologies}
\label{sec:transferability}

 \begin{figure*}
    \centering
    \subfloat[\centering Tuned (Full)]{{\includegraphics[width=0.3\linewidth]{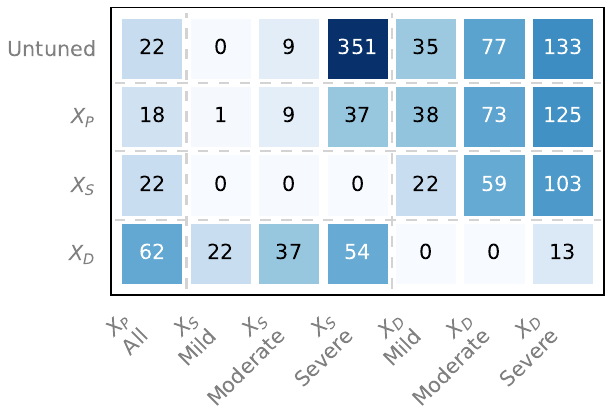} }}%
    \subfloat[\centering Tuned (Partial)]{{\includegraphics[width=0.3\linewidth]{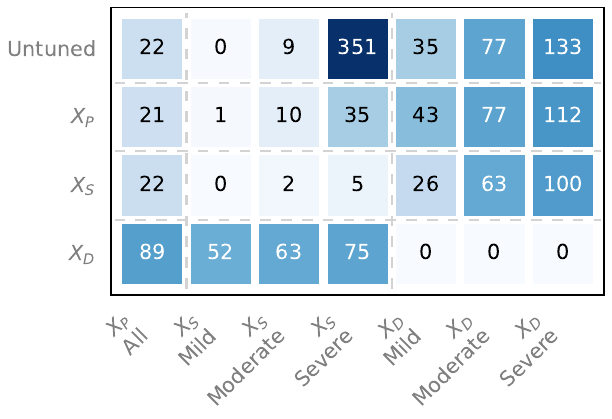} }}%
    \subfloat[\centering LoRA]{{\includegraphics[width=0.3\linewidth]{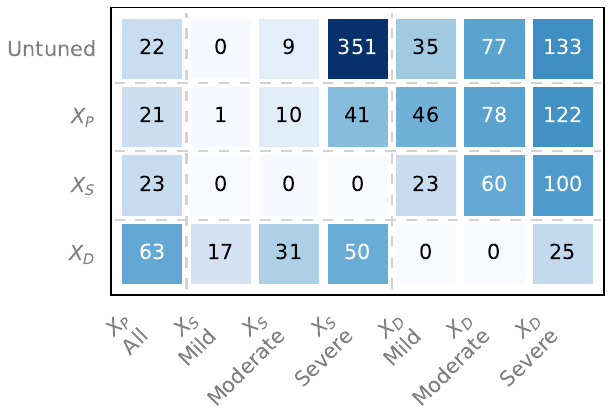} }}%
    \caption{Average speaker WER of Whisper (\texttt{tiny.en}) untuned and best fully/partially/LoRA configurations, across $X_P$ (Parkinson's), $X_S$ (stuttering) and $X_D$ (dysarthria), with various degrees of severity.}%
    \label{fig:transferability}%
\end{figure*}

Despite the state-of-the-art parameter efficiency enabled by our previous analysis, adaptations are still cohort-specific, as in prior work. We next investigate the level of personalization required to adapt to different speaker cohorts. 
As shown in Figure \ref{fig:transferability}, applying a model trained on one cohort to the same leads to the highest results, as expected. However, even within the same cohort, performance degrades for higher severities. While errors can be eliminated for mild and moderate cases, speakers with severe pathologies see the least benefit, even after full model fine-tuning. For dysarthric speech for instance, only fine-tuning or adapting the largest model yields error rates in a usable range.

Across datasets, we observe some transferability, as training on any type of atypical speech seems to improve performance on other types at least marginally. Training on $X_P$ and $X_S$ appears to transfer slightly better to each other and to $X_D$ than vice-versa. This could be an effect of the mild and moderate cases of stuttering not differing as strongly phonetically from typical speech as dysarthria. 
Also, LoRA appears to allow for more stable transferability across different atypical speech types, while preserving original performance, as shown especially for the model adapted to $X_D$. This may be because $X_D$ consists of a small vocabulary with repetitive utterances, making it prone to over-fitting when full-rank fine-tuning, whereas LoRA provides some regularization via the smaller number of adaptable parameters. 

The detailed separation of severities across these transfer results also provides indication of these methods' performance on typical speech from the same domains: Mild stuttering ($X_{S, mild}$) contains only few dysfluencies and typical pronunciation compared to $X_D$ or $X_P$. It is also the category with the consistently lowest WERs across training data regimens, including the untuned model (corresponding to Whisper's state-of-the-art transcription performance on typical speech at its time of publication; \citealp{radford2023whisper}).
Nonetheless, our experiments demonstrate the need for finer-grained personalization, as populations with severe pathologies still see the least benefits from personalization, even within their own speech disorder cohort.

\section{Dynamic Personalization}\label{sec:dynamic-personalization}
Our previous findings generalize across speech disorders to support higher parameter efficiency than prior work. However, in practice, cohort-level personalization still requires knowledge of the target etiology, as a model trained on data from one cohort does not transfer well to another. 
Additionally, severe cases of atypical speech, being rarer within the cohort, see less improvement than mild and moderate cases. Therefore, a data-centric design that is more cognizant of both speech disorder type and severity is necessary for successful personalization. Towards this goal, our second contribution is the design and use of light-weight hypernetworks \citep{ha2017hypernetworks} to dynamically generate personalized adaptation weights---essentially generating a new adapted model for each utterance at inference time. 

\subsection{Hypernetworks}

We propose that the hypernetwork is a function $H(s, c; \theta)$ with trainable parameters $\theta$ and inputs consisting of a speaker characteristics vector $s$ and the context of the generation $c$, such as the parameter type being adapted, as well as its location within the model. Both $s$ and $c$ can be manually defined (e.g., user self-identification, expert heuristics), based on external pre-trained models (e.g., speaker encoders), and/or acquired jointly during downstream meta-adaptation. The output of $H(s, c; \theta)$ are the vectorized LoRA-$A$ and $B$ matrices, which are reshaped and applied to the pre-trained weight matrix $W$ following Equation \ref{eq:lora}. In our experiments, we explore two functional forms of $H(s, c; \theta)$, namely a linear system, and an MLP with one hidden layer and ReLU activations. 

\begin{figure}
    \includegraphics[width=\linewidth]{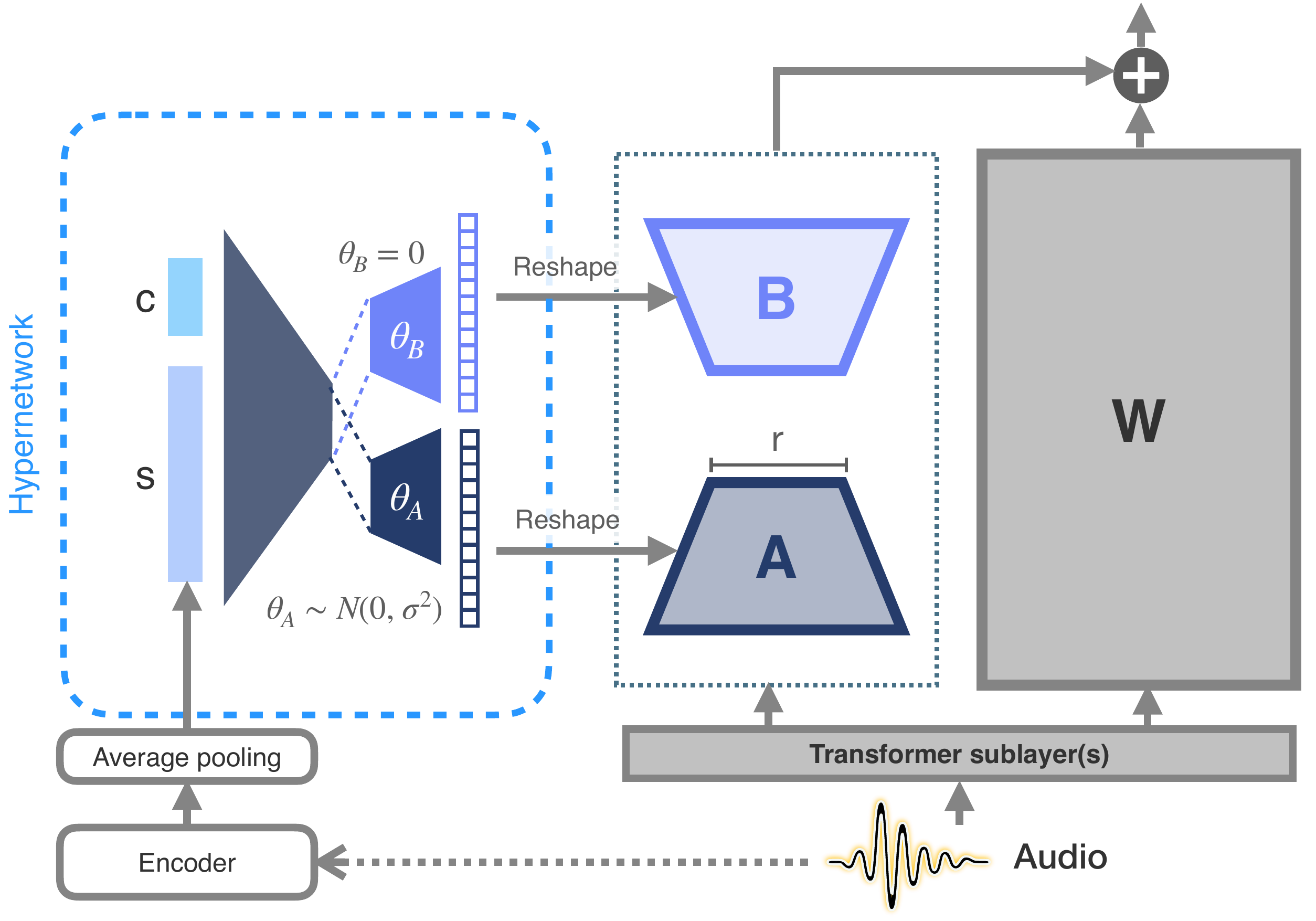}
    \caption{Hypernetwork for adapting $W$ with LoRA weights $A$ and $B$, generated by $\theta_A$ and $\theta_B$, respectively initialized with $\mathcal{N}(0, \sigma^2)$ and zeroes. Generation is conditioned on speaker characteristics $s$ from an audio encoder, and generation context $c$ denoting the target parameter's location. All trainable parameters are within the hypernetwork. }
    \label{fig:hypernetworksys}
\end{figure}

Figure \ref{fig:hypernetworksys} shows the proposed architecture to adapt a parameter $W$, where the speaker characteristics vector $s$ is computed using a speech encoder model. In our design, $s$ is computed on the utterance-level, although it is possible to replace this with a coarser speaker-level characterisation. Additionally, $H(s, c; \theta)$ consists of separate output heads for predicting $A$ and $B$, whose weights are respectively denoted by $\theta_A$ and $\theta_B$, while the remainder of the hypernetwork is shared.

\subsection{Hypernetwork Initialization}\label{sec:initialization}

When training the hypernetwork, we recommend to not trivially initialize it randomly, since the generated adaptations will equate to random pertubations that are detrimental to any existing model capabilities. For language modeling, \citet{phang2023hypertuning} propose an additional hyper-pre-training phase to first learn a hypernetwork initialization matching the host model's parameter space. However, this approach is resource intensive, requiring over 50k additional training steps, and cannot be trivially applied to ASR in a similar self-supervised manner.

Instead, we propose a simpler approach, which more closely follows the original LoRA design: specifically, initial adaptation weights, which leave the model unaugmented. For our hypernetworks, we propose implementing this design by initializing $\theta_B$ at zero, thereby nulling out any changes brought about by the initial $\Delta W$. Simultaneously, $\theta_A$ is initialized close to zero, but randomly, ensuring gradient flow during back-propagation. We found this design choice to be crucial for training as it enables learning solutions that initially match the target model's parameter space without catastrophically deteriorating performance with random noise.

\subsection{Speaker Characterization}\label{sec:characterization}

As the speaker characteristics $s$ must encode all necessary information for the hypernetwork to generate effective adaptation weights for personalization, we studied different audio-based encoding strategies to identify which factors are crucial to downstream performance. While it is possible to use manual features, such as flags to indicate speaker characteristics, we use automatic, pre-trained speech encoder models which do not require expert annotations.

In our explorations, we ablate $s$ from lower-level acoustic to higher-level concepts, such as speaker identity, leveraging speech encoder models either trained for ASR or speaker verification, respectively denoted as $s_{\mathrm{ASR}}$ and $s_{\mathrm{SV}}$. For $s_{\mathrm{ASR}}$, we use different layers from the encoder of Whisper \texttt{tiny}, \texttt{tiny.en} and \texttt{large-v2}, while for $s_{\mathrm{SV}}$, we used a speaker verification model from the Speech Brain project~\citep{speechbrain}. The speaker verification model uses an Emphasized Channel Attention, Propagation and Aggregation Time Delay Neural Network (ECAPA-TDNN; ~\citealp{Desplanques_2020}), and is trained on the VoxCeleb datasets~\citep{Nagrani17, Chung18b, Nagrani19}.

\begin{figure}
    \centering
    \subfloat[\centering Whisper ]{{\includegraphics[width=0.5\linewidth]{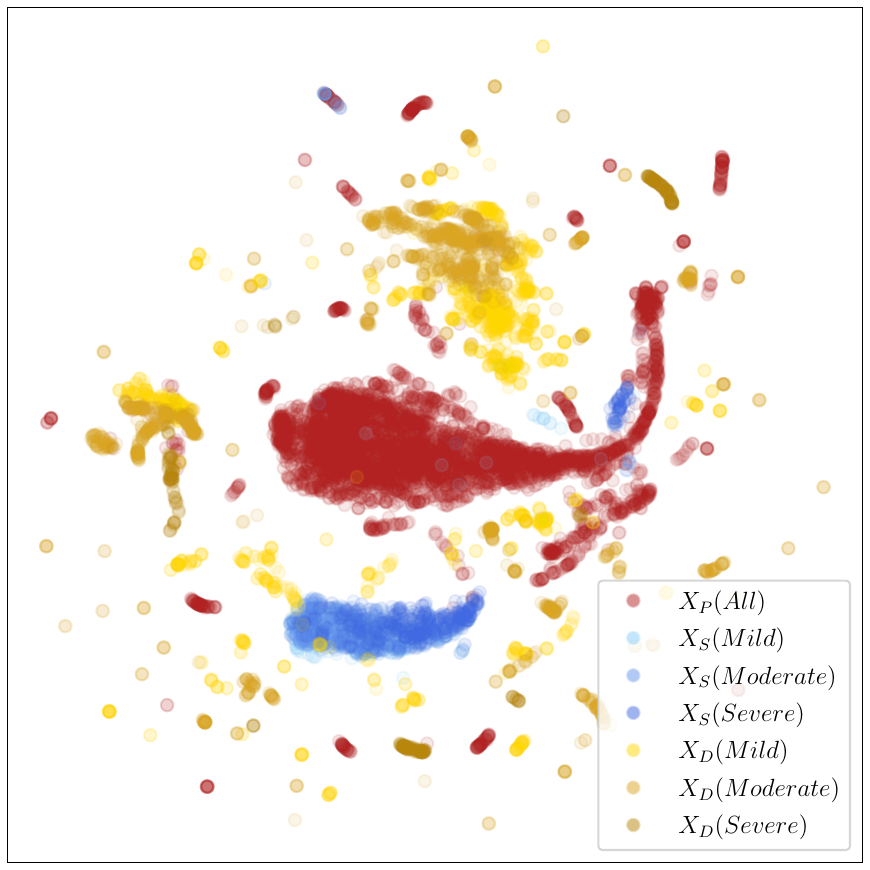} }}%
    \subfloat[\centering ECAPA-TDNN ]{{\includegraphics[width=0.5\linewidth]{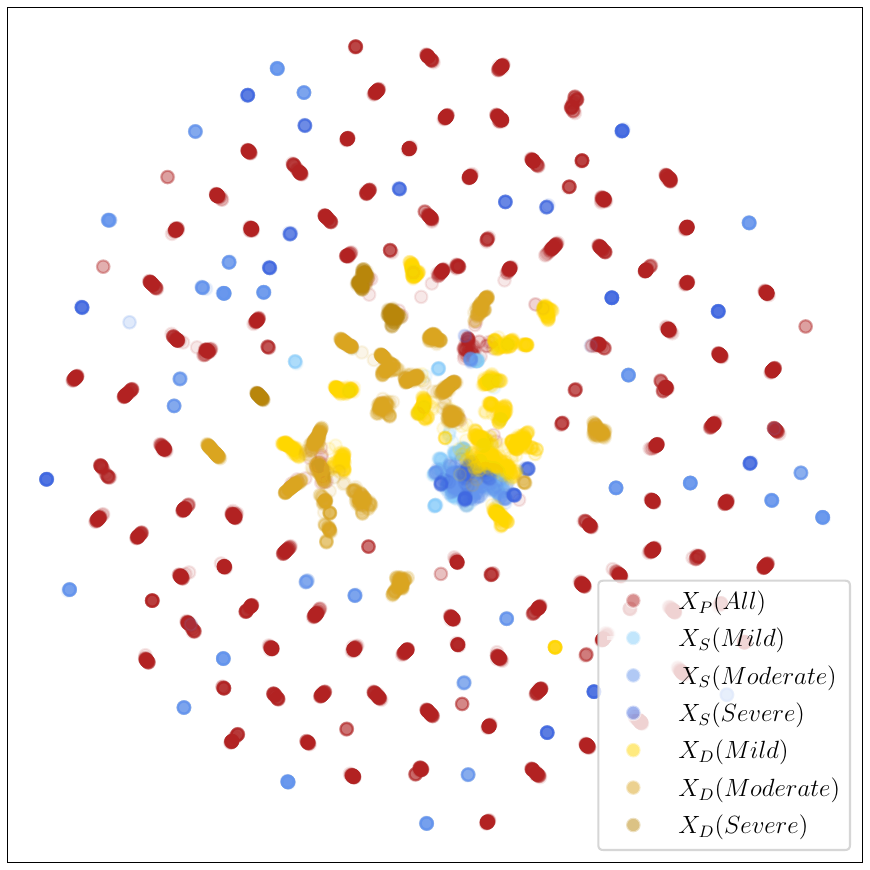} }}%
    \caption{UMAP \citep{ghojogh2023umap} visualization of speaker characterization vectors $s$ from the last encoder layer of Whisper (\texttt{large.v2}), and from the ECAPA-TDNN speaker verification model. }%
    \label{fig:speakercharacteristics}%
\end{figure}

As shown in Figure \ref{fig:speakercharacteristics}, we observe that the $s_{\mathrm{ASR}}$  embeddings are localized by etiology more strongly than  $s_{\mathrm{SV}}$. The continuity of the embeddings manifold with regards to etiology also seems to be a benefit when training the hypernetwork as we had success with $s_{\mathrm{ASR}}$ but not with $s_{\mathrm{SV}}$.

\begin{figure*}
    \centering
    \subfloat[\centering Layer 0]{{\includegraphics[width=0.23\linewidth]{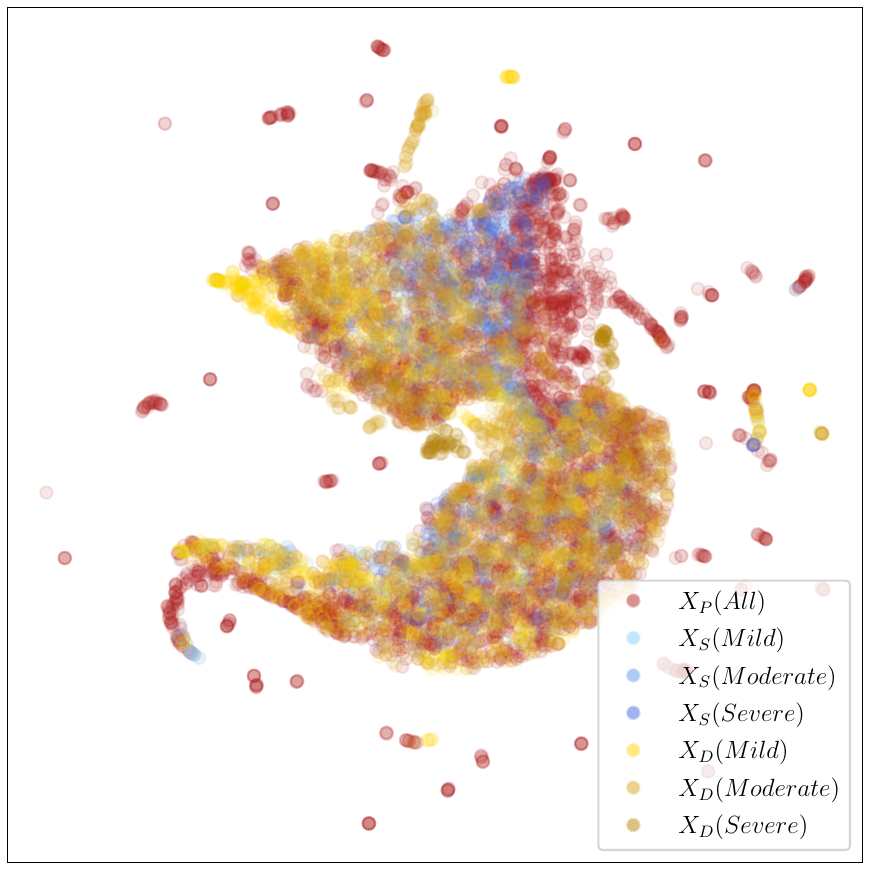} }}%
    \subfloat[\centering Layer 7]{{\includegraphics[width=0.23\linewidth]{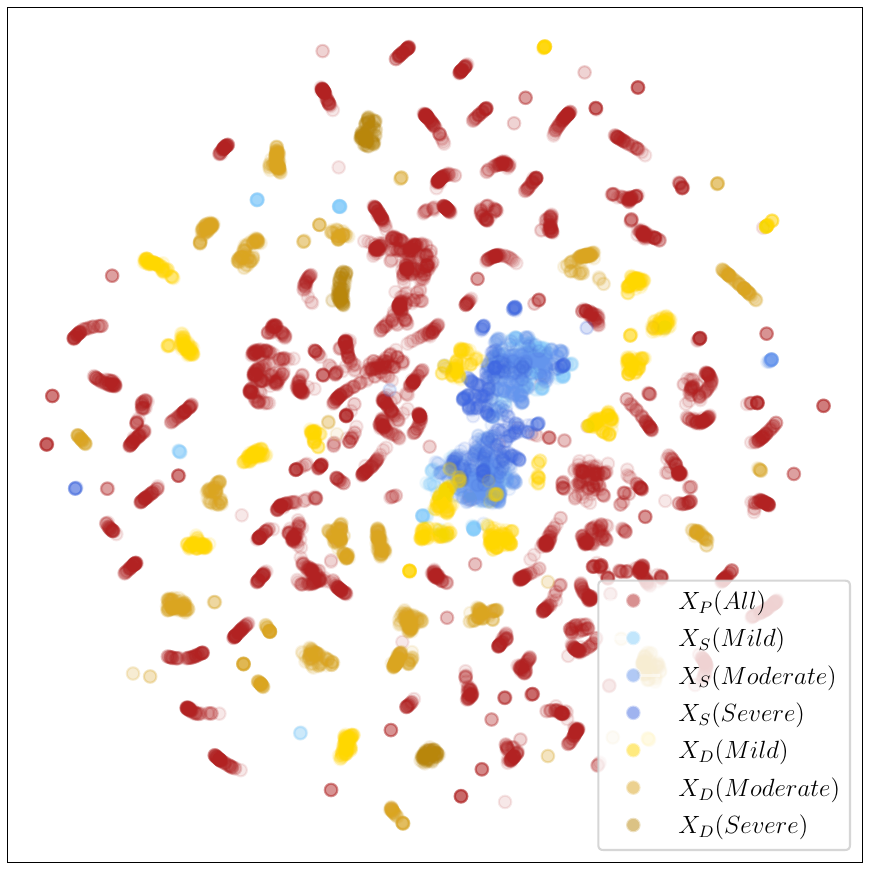} }}%
    \subfloat[\centering Layer 15]{{\includegraphics[width=0.23\linewidth]{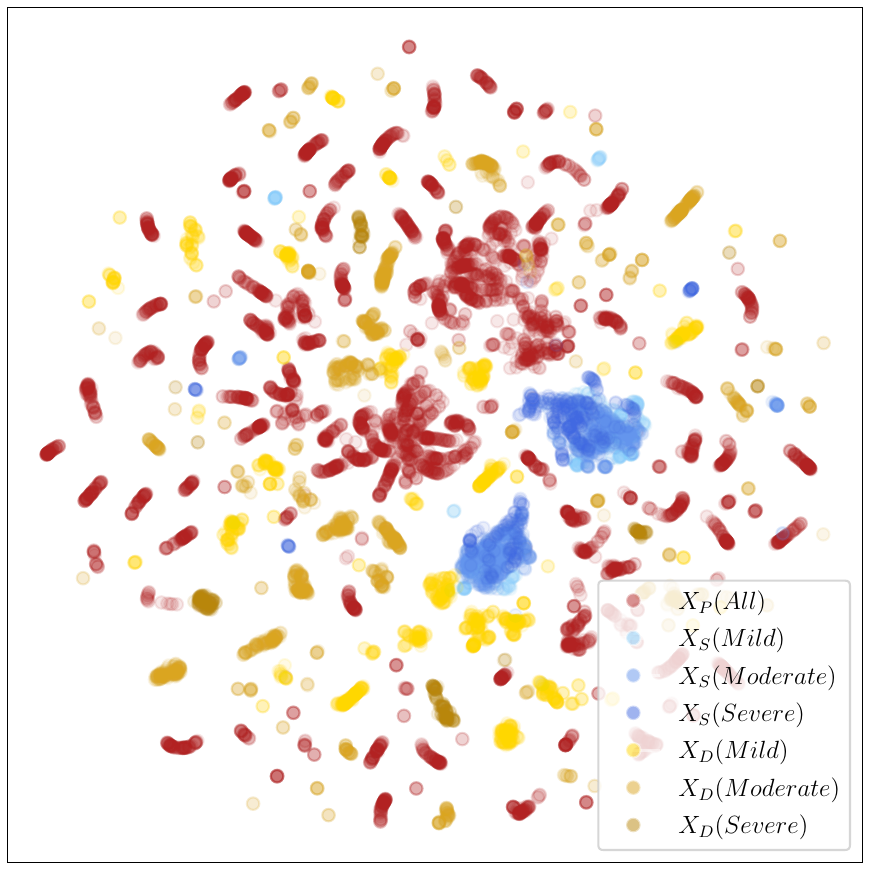} }}%
    \subfloat[\centering Layer 23]{{\includegraphics[width=0.23\linewidth]{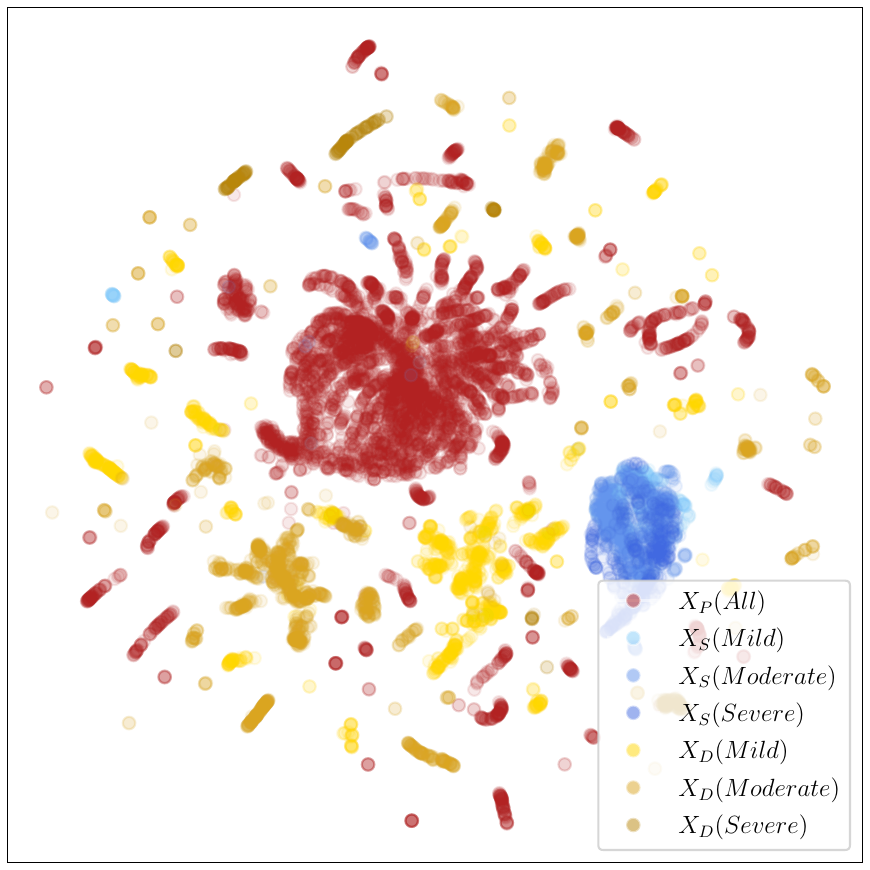} }}%
    \caption{UMAP \citep{ghojogh2023umap} visualization of Whisper (\texttt{large.v2}) encoder embeddings from various layer depths. Each point corresponds to an utterance from $X_S$ (stuttering), $X_D$ (dysarthria) and $X_P$ (Parkinson's), with color and hue corresponding to the dataset and severity of the speaker (severity annotations unavailable for $X_P$).}%
    \label{fig:encoderdepth}%
\end{figure*}

Additional to the learning task for which the speech encoder is trained on, we further explored the expressiveness of $s_{\mathrm{ASR}}$ when computed using earlier or later layers of the encoder. As shown in Figure \ref{fig:encoderdepth}, the clustering of etiology becomes more apparent in the later layers of the encoder.

Overall, we observe that effective, utterance-level adaptations require the hypernetwork to be conditioned on speaker characteristics $s$, which cover a continuous space with respect to a diverse set of features such as speaker characteristics and sufficient expressiveness of part-word acoustical units. The expressiveness of $s$ dictates whether different parameters can be generated to accommodate various speech disorders and severities thereof. The task of speaker verification, while encoding individuals distinctly, does not appear to benefit sharing of lower-level acoustic characteristics. ASR on the other hand encodes acoustic properties more continuously, and $s_{\mathrm{ASR}}$ is most effective, when higher-level features correlated with etiology begin to be encoded as well (i.e., in deeper layers).

\subsection{Final Adaptation Architecture}

Based on our findings for improving parameter efficiency (Section \ref{sec:cohort-personalization}) and hypernetwork construction (Section \ref{sec:dynamic-personalization}), our final dynamic adaptation architecture $H(s, c; \theta)$ is built as follows: A linear/MLP-based hypernetwork, which generates adaptations for the $W_1$ parameter type with rank $r =$ 2. Since the hypernetwork is shared for all instances of this parameter, $c$ provides context for the location of adaptation within the model in the form of a one-hot embedding lookup that is learned jointly during adaptation. The speaker characterization stems from the final encoder layer of Whisper \texttt{large-v2}, and is mean-pooled over time. For each forward pass, $H$ generates all adaptations for a given utterance and inserts them into the model dynamically.

\begin{table*}[thb]
    \centering
    \caption{Average speaker WER of untuned, fully/partially-tuned and LoRA/hypernetwork-adapted Whisper (\texttt{large-v2}) on test splits of $X_P$ (Parkinson's), $X_S$ (stuttering) and $X_D$ (dysarthria), reported as `P50 {\small(IQR)}’. Best fully tuned setups in \textbf{bold}, and best PEFT setups in \textbf{\textit{bolded-italics}}. Models were trained on the \textit{global} concatenation of datasets $X_{P+S+D}$, or on each individual target \textit{cohort}. On the public benchmark $X_\mathbb{P}$, we further report cohort-level adaptation results, including for \textsc{Hyper}$_\mathbb{P}$ solely trained on $X_\mathbb{P}$. Note that models trained on $X_P$ are not evaluated on $X_\mathbb{P}$ (and vice-versa), due to speaker overlap. Percentages indicate the amount of trainable parameters with respect to full fine-tuning.}
    \label{tab:hyperbenchmark}
    \resizebox{.99\textwidth}{!}{
    \begin{tabular}{@{}p{0.2cm}p{0.5cm}ccccccccc@{}}
        \toprule
        \multirow{3}{*}[-5pt]{\textsc{Setup}} & 
        & \textsc{Untuned} & 
        \multicolumn{2}{c}{\textsc{Full Tuning}}& \multicolumn{2}{c}{\textsc{Partial Tuning}} & \multicolumn{2}{c}{\textsc{LoRA}}  & \textsc{Hyper} &  \textsc{Hyper}$_{\mathbb{P}}$ \\
        \cmidrule(lr){3-3}
        \cmidrule(lr){4-5}
        \cmidrule(lr){6-7}
        \cmidrule(lr){8-9}
        \cmidrule(lr){10-10}
        \cmidrule(lr){11-11}
        &  & - & \textsc{Global} & \textsc{Cohort} & \textsc{Global}   & \textsc{Cohort} & \textsc{Global} & \textsc{Cohort} & \textsc{Global}  & \textsc{Cohort}   \\
        &   &  {\small0\%} & {\small100\%} & {\small100\%} & {\small72\%} & {\small72\%} & {\small3\%} & {\small3\%} & {\small0.1\%}  & {\small0.1\%}\\
    \midrule
    \addlinespace[.5em]
          $X_{P}$ & & 8.5 {\small(17.7)} &	\textbf{4.4 {\small(9.4)}} &	5.2 {\small(9.4)} &	5.0 {\small(9.8)}  &	 5.0 {\small(10.1)} & 6.9 {\small(14.9)} &	7.1 {\small(16.3)} &	\textbf{\textit{6.0 {\small(13.8)}}}  &	- \\
        \midrule
           $X_{S}$ & &  19.3 {\small(171.0)} &	0.0 {\small(1.9)} &	\textbf{0.0 {\small(1.0)}} &	0.0 {\small(2.5)}  &	0.0 {\small(1.5)} &	0.2  {\small(6.2)} &	\textbf{\textit{0.0 {\small(4.0)}}} &	\textbf{\textit{0.0 {\small(4.0)}}} & 0.0 {\small(8.9)}\\
           $X_{S, mild}$ & &  0.0 {\small(0.9)} &	\textbf{0.0 {\small(0.0)}} &	\textbf{0.0 {\small(0.0)}} &	\textbf{0.0 {\small(0.0)}}  &	\textbf{0.0 {\small(0.0)}} &	0.0  {\small(22.9)} &		\textbf{\textit{0.0 {\small(0.0)}}} &		\textbf{\textit{0.0 {\small(0.0)}}} &	0.0 {\small(0.0)}\\
           $X_{S, mod}$&  & 1.5 {\small(12.4)} &	0.0 {\small(1.6)} &	\textbf{0.0 {\small(0.7)}} &	0.0 {\small(2.0)} &	\textbf{0.0 {\small(0.7)}} &	0.4 {\small(4.4)} &	0.0 {\small(3.6)} &	\textbf{\textit{0.0 {\small(2.9)}}} &	0.0 {\small(6.4)} \\
           $X_{S, sev}$  & & 70.6 {\small(626.6)} &	0.5 {\small(4.4)} &	\textbf{0.0 {\small(2.6)}} &	0.0 {\small(6.0)}  &	0.0 {\small(4.5)} &	0.0  {\small(15.7)} &	\textbf{\textit{0.0 {\small(8.8)}}} &	0.0 {\small(10.0)} &	0.0 {\small(22.3)}\\
           \midrule
           $X_{D}$ &   & 33.3 {\small(63.8)} &	0.0 {\small(9.5)} &	\textbf{0.0 {\small(2.4)}} &	0.0 {\small(9.5)} & 0.0 {\small(5.3)}&	15.3 {\small(35.1)} &	\textbf{\textit{7.1 {\small(14.7)}}} &	8.3 {\small(8.9)} &	24.3 {\small(54.6)}\\
           $X_{D, mild}$ &   & 0.0 {\small(38.9)} & \textbf{0.0 {\small(0.0)}} & \textbf{0.0 {\small(0.0)}} & \textbf{0.0 {\small(0.0)}} &	\textbf{0.0 {\small(0.0)}}  &	0.0 {\small(6.5)} &	\textbf{\textit{0.0 {\small(0.0)}}} &	\textbf{\textit{0.0 {\small(0.0)}}} &	0.00 {\small(27.8)} \\
           $X_{D, mod}$ &   & 44.4 {\small(87.4)} &	\textbf{0.0 {\small(0.0)}} &	\textbf{0.0 {\small(0.0)}} &	\textbf{0.0 {\small(0.0)}} &	\textbf{0.0 {\small(0.0)}} &	7.87 {\small(49.4)} &	0.0 {\small(10.8)} &	\textbf{\textit{0.0 {\small(0.0)}}} &	25.5 {\small(74.1)}\\
           $X_{D, sev}$ &   & 100.0 {\small(68.1)} &	0.0 {\small(66.7)} &	\textbf{0.0 {\small(16.7)}} &	0.0 {\small(66.7)} &	0.0 {\small(37.2)} &	83.3 {\small(77.8)} &	\textbf{\textit{50.0 {\small(70.6)}}} &	58.3 {\small(62.5)} &	93.3 {\small(76.4)}\\
           \midrule
           $X_{P+S+D}$ &   & 16.2 {\small(83.7)} &	\textbf{2.1 {\small(6.5)}} &	2.4 {\small(5.2)} &	2.4 {\small(6.9)} &	2.3 {\small(6.1)} &	5.5 (14.41 &	4.4 {\small(11.3)} & \textbf{\textit{4.0 {\small(9.3)}}} &	- \\
           \midrule
           $X_{\mathbb{P}}$ &   & 11.8 {\small(33.6)} &	- &	\textbf{1.4 {\small(7.7)}} &	- &	2.5 {\small(10.0)} &	-  &	\textbf{\textit{3.5 {\small(16.3)}}} & - &	4.1 {\small(17.1)}\\
    
        \bottomrule
    \end{tabular}
}
\end{table*}

\begin{table}[tbh]
    \centering
    \caption{Average speaker WER of individual-level personalized Whisper models on test splits of $X_P$, $X_S$, and $X_D$, reported as `P50 {\small(IQR)}'. Best fully tuned setups in \textbf{bold}, and best PEFT setups in \textbf{\textit{bolded-italics}}. Full/partial/low-rank adaptations are trained using 70\% of an individual's data, while the hypernetwork generates zero-shot adaptations without any data from the individual. Fine-tuned setups are initialized using the corresponding cohort model. Percentages indicate the amount of trainable parameters with respect to full tuning. 
    } 
    \label{tab:individual_ablations}
    \resizebox{\columnwidth}{!}{
    \begin{tabular}{@{}p{1.5cm}cccc@{}}
        \toprule
        \multirow{2}{*}[-2pt]{\textsc{Setup}} 
    &  \textsc{Full} &  \textsc{Partial}  & \textsc{LoRA} &  \textsc{Hyper}   \\
    &   {\small100\%} & {\small72\%} & {\small3\%} & {\small0.1\%} \\
        \midrule
         {$X_{P}$} &  4.7 {\small(9.3)} &	\textbf{4.4 {\small(9.3)}} &	8.1 {\small(16.8)} &	\textbf{\textit{7.1 {\small(23.9)}}} \\
        \midrule
        $X_{S}$  &  \textbf{0.0 {\small(1.5)}} &	0.0 {\small(1.9)} &	1.4 {\small(12.2)} &	 \textbf{\textit{0.0 {\small(5.1)}}} \\
         \midrule
        $X_{D}$  & \textbf{0.0 {\small(1.1)}} &	0.0 {\small(4.4)} &	\textbf{\textit{0.0 {\small(6.7)}}} &	 7.0 {\small(7.9)}  \\
          \midrule
        $X_{P+S+D}$  &  2.2 {\small(5.1)} &	\textbf{2.1 {\small(5.8)}} & 4.4 {\small(13.7)} &	 \textbf{\textit{4.3 {\small(14.3)}}}  \\
        \bottomrule
    \end{tabular}
    }
\end{table}

\section{Results}\label{sec:results}
We next compare our proposed approach to the \textit{full}, \textit{partial} and \textit{LoRA} baselines outlined in Section \ref{sec:setup}, and report the findings in Tables ~\ref{tab:hyperbenchmark} and ~\ref{tab:individual_ablations} in decreasing order of heterogeneity of the training data, namely at the \textit{global}, \textit{cohort} and \textit{individual} level. Results are reported for Whisper \texttt{large-v2}, which represents the upper bound in terms of performance. 

\subsection{Global Adaptation}\label{sec:global-adaptation}
From Table \ref{tab:hyperbenchmark}, we observe that global adaptation, i.e., training on data from all cohorts simultaneously, works well for people with mild to moderate speech differences. Indeed, the speaker-wise median WER of 0 reflects our initial observation from Section \ref{sec:cohort-personalization} in that the majority of common voice commands are covered well using most adaptation approaches. As mentioned in Section \ref{sec:transferability}, $X_{S, mild}$ further indicates that all approaches would likely perform comparably to the untuned model on typical speech. Even on $X_P$, which has the most diverse set of utterances, the WER and IQR typically fall within the 15\% usability threshold. This global approach further circumvents the need of managing cohort-specific models, however some cohorts are more represented than others, leading to model bias against rarely observed cohort characteristics. For example, we generally see higher WERs for higher severities, which are more rarely observed, and notably poorer performance for severe dysarthria $X_{D, sev}$ with only two speakers seen during training. These effects are most prominent when the tunable parameter budget is low, as with LoRA. Given the ability to tune larger parts of the model, i.e., full and partial fine-tuning, the globally tuned model can still be applied to a broader range of atypical speech types, however this requires tuning 72\%--100\% of the 1.6B parameters. Our proposed approach of using hypernetworks to dynamically generate personalized adaptations appears to most effectively leverage global data sharing, outperforming standard LoRA and even full fine-tuning on $X_{S, sev}$, for an overall WER of 4.0, while using 0.1\% of the full parameter budget. It further maintains the base model's original performance on close-to-typical speech ($X_{S, mild}$) best, as indicated by its substantially lower IQR compared to LoRA.

\subsection{Cohort-level Adaptation}\label{sec:cohort-adaptation}

When knowledge of a speaker's cohort-membership is available, we observe from Table \ref{tab:hyperbenchmark} that fine-tuning on cohort-specific data provides better performance than global adaptation regardless of the fine-tuning technique applied, alluding to the need for a higher degree of personalization. However, this increased granularity comes at the cost of necessitating one model per atypical speech type. 

In contrast, we find that despite having to share a single model across all cohorts, lacking explicit knowledge of a target speaker's etiology, and representing a magnitudes smaller architecture, hypernetworks are able to generate adaptations which are competitive to cohort-level full fine-tuning, and LoRA. This improvement in performance could be attributed to the relatively flexible inductive bias imposed on the hypernetwork, allowing it to share representations that may be beneficial across heterogeneous cohorts. The cross-cohort transfer of \textsc{Hyper}$_\mathbb{P}$ to $X_{S}$ and $X_{D}$ reflects this capability in particular, as it has substantially lower WER than the untuned model, despite only being trained on the $X_\mathbb{P}$ cohort. As the global hypernetwork, trained on $X_{P+S+D}$, nonetheless outperforms the cohort variant, training on a diverse set of speech characteristics appears crucial for learning sharable representations. We further examine this hypothesis in Section \ref{sec:analysis-parameters}.

\subsection{Individualized Adaptation}\label{sec:individual-adaptation}
Moving to the highest level of personalization, we next compare our zero-shot hypernetworks to full/partial/LoRA-tuned, individually personalized ASR models in Table~\ref{tab:individual_ablations}. For the baselines, 70\% of the data for each test subject in Table~\ref{tab:hyperbenchmark} is used to continually fine-tune their relevant cohort-level model, while for the hypernetwork, we evaluate on the same 30\% remainder of the test data, but do not train on any target speaker data. Similarly to prior work utilizing target speaker data---either via continual fine-tuning \citep{tomanek2021ondevice} or via the fusion of multiple individualized models based on their similarity to the target speaker \citep{qi2023parameterefficient}---this individual-level personalization generally improves the baselines' performance. However these approaches require training, retaining and selecting an even higher number of models than at the cohort-level. Conversely, the adaptations generated by our single hypernetwork remain competitive, especially to individualized LoRA, despite not observing any training data from the target speaker and having no information regarding their cohort-membership. Relative to the performance upper-bound of full and partial fine-tuning, the hypernetwork provides 2\% higher WER, however remaining far below the usability threshold of 15\% WER for most speakers, while requiring up to three orders of magnitude fewer tuned parameters.

\subsection{Analysis of Parameter Space}\label{sec:analysis-parameters}
\begin{figure}
    \centering
    \subfloat[\centering Linear]{{\includegraphics[width=0.5\linewidth]{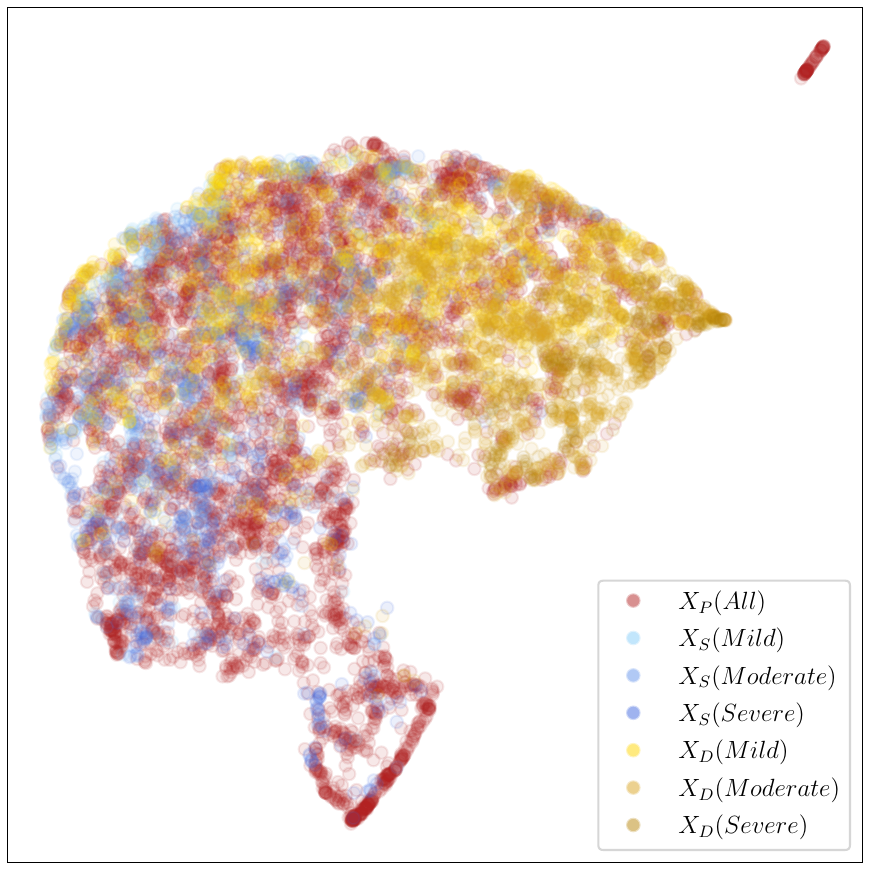} }}%
    \subfloat[\centering MLP ]{{\includegraphics[width=0.5\linewidth]{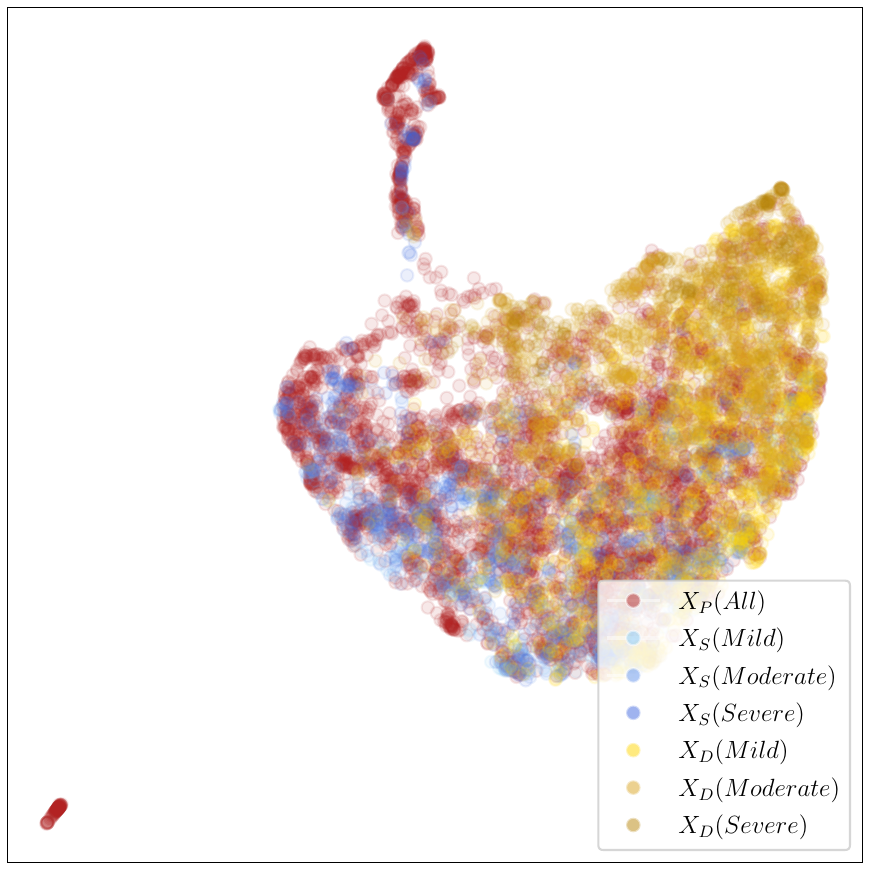} }}%
    \caption{UMAP \citep{ghojogh2023umap} visualization of adaptions $W'$ for 10k utterances from $X_S$ (stuttering), $X_D$ (dysarthria) and $X_P$ (Parkinson's) generated by a linear or MLP-based hypernetwork, colored by dataset and severity.}%
    \label{fig:hypernetworkarchablation}%
\end{figure}

To gain a better understanding of how hypernetworks balance global knowledge sharing while maintaining effective individualized personalization, we perform an analysis of the generated parameter manifold with respect to the different atypical speech cohorts being represented. Figure \ref{fig:hypernetworkarchablation} shows a subsample of generated parameters for 10k utterances in the test set. Regardless of the hypernetwork's functional form, there are regions where generated adaptations overlap across all three datasets. We also observe the general trend that there is some overlap between dysfluent utterances $X_S$ and those associated with Parkinson's $X_P$, while there is far less sharing between $X_P$ and $X_D$. This aligns with the previous observations from Table \ref{tab:hyperbenchmark}, where the hypernetwork trained solely on Parkinson's data transfered well to stuttering, but not to speech consistent with Cerebal Palsy. As these overlaps do not occur in the original speech characterizations $s$ (Figures \ref{fig:speakercharacteristics} and \ref{fig:encoderdepth}), we stipulate that the hypernetwork was able to learn common adaptations that generalize well cross different speech disorders, while simultaneously generating adaptations that are unique to an etiology as seen in the non-overlapping regions.

In terms of the functional form of $H$, we observe only minor performance differences between the linear and MLP variant, with the latter exhibiting slightly lower WER overall. Together with the similar overlaps exhibited by the generated parameters, this points towards the general meta-learned individualization and knowledge sharing approach being more crucial than the exact form of the architecture employed to achieve this goal.

\subsection{Alternative Backbone Architectures}\label{sec:alternative-backbones}

\begin{table}[thb]
    \centering
    \caption{Average speaker WER of LoRA and hypernetwork trained on $X_{P+S+D}$, applied to Whisper \texttt{tiny}/\texttt{tiny.en}, and evaluated on test splits of $X_P$, $X_S$, and $X_D$, reported as `P50 {\small(IQR)}', with best setups in \textbf{bold}. Percentages indicate the amount of tuned parameters with respect to full fine-tuning.}
    \label{tab:alternativebackbone}
    \resizebox{.99\linewidth}{!}{
    \begin{tabular}{@{}p{1cm}ccccc@{}}
        \toprule
        \multirow{3}{*}[-2pt]{\textsc{Setup}}  & \multicolumn{2}{c}{\texttt{tiny}} &
        \multicolumn{2}{c}{\texttt{tiny.en}} \\ 
        \cmidrule(lr){2-3}
        \cmidrule(lr){4-5}
        & \textsc{LoRA} & \textsc{Hyper} & \textsc{LoRA}   & \textsc{Hyper} \\ 
        & \small{0.08\%} & \small{1.47\%} & \small{0.08\%} & \small{1.47\%}  \\ 
        \midrule
        $X_{P}$ & 	21.7 {\small(36.0)}  & \textbf{20.4 {\small(35.9)}}	 & 20.1 {\small(31.7)} & \textbf{20.1 {\small(31.2)}}\\ 
        $X_{D}$ & 	42.0 {\small(69.5)}  & \textbf{28.4 {\small(49.5)}} & 24.5 {\small(56.5)}	 & \textbf{19.7 {\small(36.4)}}\\ 
        $X_{S}$  &	7.0 {\small(24.7)}  & \textbf{5.4 {\small(22.0)}} & 6.7 {\small(21.5)} & \textbf{5.1 {\small(21.8)}}	\\ 
        \midrule
        $X_{P+S+D}$ & 18.9 {\small(36.4)}	  & \textbf{15.7 {\small(32.4)}} & 15.5  {\small(31.3)}	 & \textbf{14.2 {\small(28.3)}}\\ 
    \bottomrule
    \end{tabular}
}
\end{table}

So far, we have found hypernetworks to consistently generate effective adaptations compared to more training-intensive approaches, across global, cohort and individual-level personalization as heterogeneity of atypical speech characteristics increase (Tables \ref{tab:hyperbenchmark} and \ref{tab:individual_ablations}). To explore how generalizable our findings are with respect to alternative backbone architectures, we next apply hypernetworks to host models with different architectures and/or pre-training data, namely Whisper \texttt{tiny} and \texttt{tiny.en} respectively. Both consist of 4 instead of \texttt{large-v2}'s 32 encoder/decoder layers, having 0.02\% the size of the larger architecture, with \texttt{tiny.en} having further been trained on English speech alone. Based on the findings from Figure~\ref{fig:ssa-heatmap}, which indicate that $W_1$ accounts for the largest adaptations, we focus on adapting only the $W_1$ matrices of the encoder and decoder of the \texttt{tiny/tiny.en} models.

Table \ref{tab:alternativebackbone} shows that hypernetworks continue to consistently outperform LoRA across all datasets, despite being trained just once overall, instead of once per-cohort, and having a substantially smaller parameter budget. Similarly to our initial experiments in Section \ref{sec:cohort-personalization}, we observe slightly higher decoding stability and fewer hallucinations for the monolingual \texttt{tiny.en} model. In general, the smaller backbones have a lower base performance compared to \texttt{large-v2}, however relative to untuned \texttt{tiny.en}, we are nonetheless able to reduce WERs from 22.4 $\rightarrow$ 20.1 for $X_P$, 97.3 $\rightarrow$ 5.1 for $X_S$, and 67.4 $\rightarrow$ 19.7 for $X_D$, using hypernetworks-generated LoRA. With an average WER of 14.2 across datasets, hypernetworks further bring us into the range of practical usability, despite the much more parameter-constrained environment. The overall approach of using hypernetworks to dynamically generate adaptation parameters therefore seems robust to not only different types of atypical speech (Sections \ref{sec:global-adaptation}, \ref{sec:cohort-adaptation} and \ref{sec:individual-adaptation}), and choice of parameter generator (Section \ref{sec:analysis-parameters}), but also with respect to the host model architecture.

\section{Conclusion}

Due to data scarcity and the highly variable nature of atypical speech, prior work relied on cohort-level fine-tuning and PEFT, followed by individualized adaptation. While this approach substantially reduces WER, it necessitates training as many models as there are speakers, and requires expert knowledge of the cohort a speaker belongs to. In Section \ref{sec:transferability}, we demonstrate that this knowledge is critical, as a model trained on one cohort does not transfer to others. In addition, higher severity speakers and those with multiple pathologies still do not benefit from the improvements for mild and moderate cases, as they are least represented in the data.

Combining meta-level knowledge sharing and highly individualized personalization, we therefore proposed using hypernetworks to dynamically generate adaptations during inference (Section \ref{sec:dynamic-personalization}). As generating an entire model is computationally infeasible, we first conducted a study regarding which individual model parameters have the highest influence on adaptation performance (Section \ref{sec:cohort-personalization}). Analyzing model components at increasing levels of detail using a novel combination of LoRA and SSAs, we identified a single parameter type $W_1$ which contributes most to adaptation performance. As this effect was consistent across different model sizes, pre-training strategies and LoRA ranks, we were able to halve WER while using 0.03\% of the parameter budget required for full fine-tuning.

Based on these findings, we were able to scale our hypernetwork-based ASR adaptation approach to an even smaller parameter budget of 0.01\%. Despite sharing a single model across cohorts and having access to neither cohort or individual speaker information, hypernetworks reach a WER of 4.0, consistently outperforming LoRA, and performing competitively to full fine-tuning (Section \ref{sec:global-adaptation}). These results hold, even when the latter two approaches are specifically fine-tuned using in-cohort data (Section \ref{sec:cohort-adaptation}) and/or training data from a target individual (Section \ref{sec:individual-adaptation}). Further ablating the hypernetworks' parameter generators (Section \ref{sec:analysis-parameters}) and host model architectures (Section \ref{sec:alternative-backbones}), we find our general meta-learning approach to generalize well across both datasets and models.

Our analyses in Sections \ref{sec:initialization}, \ref{sec:characterization} and \ref{sec:analysis-parameters} further surface factors critical to hypernetwork performance: Firstly, improving upon prior approaches, we find that nulling out initial adaptations is crucial for not deteriorating existing model performance. Second, for speaker characterization, continuous coverage across acoustic features appears more important than discrete speaker featurizations. Third, the hypernetwork meta-learns adaptation weights which exhibit overlaps, matching etiological intuitions beyond the information present in the input embeddings. Improving upon prior work, we demonstrate that this general approach holds across multiple types of atypical speech, enabling zero-shot dynamic personalization without explicit knowledge of cohort membership, nor training data from the target speaker.

In this work, we target a diverse set of dysfluency and phonology-related speech disorders. However, future work may investigate other categorizations of atypical speech that are not included in our datasets. Additionally, we believe exploring a broader range of ASR scenarios could be fruitful, especially since standard adapters have been shown to work well for heavy-accented speech, and multi-lingual settings ~\citep{le2021,tomanek2021adapters}. Applying hypernetworks to these, as well as a myriad of other downstream scenarios, could therefore form an interesting extension to this work. Lastly, our experiments centered around multiple variants of Whisper, confirming our findings across different model sizes and pre-training data regimens. We believe future work could follow our general approach for identifying relevant adaptation parameters, and subsequently generate them using meta-learned hypernetworks, in order to better optimize parameter efficiency when adapting alternative backbone architectures in low-resource scenarios.

\section*{Acknowledgements}
Thanks to  Leah Findlater, Pooja Solanki, and Jeffrey Bigham for supporting this project, as well as to the TACL action editor and the anonymous reviewers for their helpful comments and insightful discussions.

\bibliography{references}
\bibliographystyle{acl_natbib}

\appendix

\end{document}